\def\year{2021}\relax
\documentclass[letterpaper]{article} 
\usepackage{aaai21}  
\usepackage{times}  
\usepackage{helvet} 
\usepackage{courier}  
\usepackage[hyphens]{url}  
\usepackage{graphicx} 
\usepackage{graphicx} 
\usepackage{natbib}  
\usepackage{caption} 
\newcommand{\mli}[1]{\mathit{#1}}

\usepackage[switch]{lineno}

\usepackage{adjustbox}

\usepackage{amssymb}
\usepackage{amsmath, calc}
\usepackage{newtxtext,newtxmath}
\usepackage{booktabs}
\usepackage{multirow}
\usepackage{graphicx}
\usepackage{tabularx}

\graphicspath{ {./images/} }

\usepackage[]{algorithm2e}

\usepackage{tikz}
\usetikzlibrary{decorations.text,calc,shapes,arrows,arrows.meta, positioning,shapes.misc,decorations.markings,decorations.markings,decorations.pathreplacing,matrix,spy}

\usepackage{placeins}
\usepackage{wrapfig}
\usepackage{pgfplots}

\newcommand{\textcite}[1]{\citet{#1}}

\everypar{\looseness=-1} 
\newcommand{\cuML}{cuML UMAP}

\title{Bringing UMAP Closer to the Speed of Light with GPU Acceleration}

\author{
  Corey J. Nolet\textsuperscript{\rm 1,2}, 
  Victor Lafargue\textsuperscript{\rm 1}, 
  Edward Raff\textsuperscript{\rm 2,3}, 
  Thejaswi Nanditale\textsuperscript{\rm 1}, 
  Tim Oates\textsuperscript{\rm 2}, \\
  John Zedlewski\textsuperscript{\rm 1},  
  Joshua Patterson\textsuperscript{\rm 1}  \\
}

\affiliations {
    \textsuperscript{\rm 1}Nvidia, 
    \textsuperscript{\rm 2}University of Maryland Baltimore County, 
    \textsuperscript{\rm 3}Booz Allen Hamilton \\
}

\begin{document}
\maketitle

\begin{abstract}
The Uniform Manifold Approximation and Projection (UMAP) algorithm has become widely popular for its ease of use, quality of results, and support for exploratory, unsupervised, supervised, and semi-supervised learning. While many algorithms can be ported to a GPU in a simple and direct fashion, such efforts have resulted in inefficient and inaccurate versions of UMAP. We show a number of techniques that can be used to make a faster and more faithful GPU version of UMAP, and obtain speedups of up to 100x in practice. Many of these design choices/lessons are general purpose and may inform the conversion of other graph and manifold learning algorithms to use GPUs. 
Our implementation has been made publicly available as part of the open source RAPIDS cuML library (https://github.com/rapidsai/cuml).
\end{abstract}

\section{Introduction}

Like other manifold learning algorithms, the Uniform Manifold Approximation and Projection algorithm (UMAP)~\cite{mcinnes2018umap} relies upon the manifold hypothesis \cite{fefferman2016testing} to preserve local neighborhood structure by modeling high-dimensional data in a low-dimensional space. This is in contrast to linear dimensionality reduction techniques like PCA, which aim only to preserve global Euclidean structure \cite{he2005neighborhood}.
UMAP produces low-dimensional embeddings that are useful for both visual analytics and downstream machine learning tasks. Unlike other manifold learning algorithms, such as IsoMap \cite{tenenbaum2000global}, Locally Linear Embeddings (LLE) \cite{roweis2000nonlinear}, Laplacian Eigenmaps \cite{belkin2002laplacian}, and t-Distributed Stochastic Neighbor Embeddings (T-SNE) \cite{maaten2008visualizing}, UMAP has native support for supervised, unsupervised, and semi-supervised metric learning. Since its introduction in 2018, it has found use in exploratory data analysis applications \cite{ordun2020exploratory,wander2020exploratory,obermayer2020scelvis,oden2020lessons}, as well as bioinformatics, cancer research \cite{andor2018joint}, single-cell genomics\cite{travaglini2019molecular,becht2018evaluation,clarapar25:online}, and the interpretation of highly non-linear models like deep neural networks \cite{carter2019exploring}. This combination of features and quality of results has made UMAP a widely used and popular tool. 


The wide array of applications and use of UMAP makes it desirable to produce faster versions of the algorithm. This is compounded by an increasing demand for exploratory and interactive visualization \cite{carter2019exploring,pezzotti2018linear,chatzimparmpas2020t,obermayer2020scelvis}, that necesitates a lower latency in results. GPUs are a strong candidate for acheiving faster implementation by trading per-core clock speeds for signficantly more cores, provided that they can be utilized effectively in parallel computation. A direct conversion of UMAP to the GPU already exists in the GPUMAP~\cite{p3732gpu8:online} project but, due to technical details, is not always faithful in reproducing the same quality of results. In this work, we show that applying a few general techniques can produce a version that is both faster and faithful in its results.


This paper contributes three components to the growing ecosystem \cite{okuta2017cupy,raschka2020machine,johnson2019billion,paszke2019pytorch} of GPU-accelerated tools for data science in Python. First, we contribute a near drop-in replacement of the UMAP-learn Python library, which has been GPU-accelerated end-to-end in CUDA/C++. The speedup of this new implementation is evaluated against the current state-of-the-art, including UMAP-learn on the CPU and an existing GPU port of UMAP. Second, we contribute a GPU-accelerated near-drop-in replacement of the trustworthiness score, which is often used to evaluate the extent to which manifold learning algorithms preserve local neighborhood structure. Finally, we contribute a distributed version of UMAP and provide empirical evidence of its effectiveness.

The remainder of our paper is organized as follows. We will discuss related work in Section \ref{sec:related_work}, with a brief review of UMAP in Section \ref{sec:umap}. Our approach to implementing UMAP for the GPU is detailed in Section \ref{sec:gpu_acceleration}, with code available  as part of the RAPIDS cuML library (https://github.com/rapidsai/cuml). Our results show up to $100\times$ speedups in Section \ref{sec:experiments}, followed by our conclusions in Section \ref{sec:conclusion}. 

\section{Related Work} \label{sec:related_work}




AI and ML research often strikes a balance in scientific exploration and software engineering, with well-engineered software having a dramatic impact on both researchers and practioners. Well implemented single-purpose packages that contain a single method or limited scope have proven to have a large influence on both of these factors. Many such works have limited theoretical contribution, but instead detail the work that goes into a careful and thoroughly efficent version of the algorithm. LIBLINEAR~\cite{Fan2008} and LIBSVM~\cite{Chang2011} became tools for users and software to modify for years of research. Similar has happened for Optuna \cite{Akiba:2019:ONH:3292500.3330701} with bayesian hyper-parameter optimization, and XGBoost~\cite{xgboost} has played a role in re-kindling decision tree \& boosting research, its details on coding optimizations impacting other influential tools like LightGBM \cite{NIPS2017_6907} and CatBoost~\cite{CatBoost}. Likewise the original Cover-Tree implementation~\cite{Beygelzimer2006} has influenced over a decade of nearest-neighbor search. The UMAP algorithm has quickly fallen into this group of single, influential implementations, but lacks GPU acceleration. We resolve this shortcoming in our work while simultaniously improving UMAP's qualtiative results through better implementation design, providing an empirical "existance proof" that these issues are not fundamental limitations of the original algorithm. 

UMAP's ability to shortcut the need for storing $n^2$ pairwise distances by defining local neighborhoods with the k-nearest neighbors around each data point is like other manifold learning algorithms. T-SNE predates UMAP and has found popularity in many of the same communities that UMAP has now become considered the state of the art \cite{oden2020lessons}. T-SNE models point distances as probability distributions, constructing a students-t kernel from training data and minimizing the Kullback-Liebler divergence against the low-dimensional representation. While originally intractable for datasets containing more than a few thousand points, GPU-accelerated variants have recently breathed new life into the algorithm \cite{chan2018t}. Still, T-SNE has not been shown to work well for downstream machine learning tasks and lacks support for supervised learning.

The reference implementation of UMAP is built on top of the Numba\cite{lam2015numba} library and uses just-in-time (JIT) compilation to make use of parallel low-level CPU optimizations. The GPUMAP library~\cite{p3732gpu8:online}
is a direct port of the reference library to the GPU, using Numba's \texttt{cuda.jit} feature, along with the CuPy library, to directly replace many SciPy library invocations with CUDA-backed implementations. Like other libraries that require fast nearest neighbors search on GPUs \cite{chan2018t}, GPUMAP uses the FAISS library~\cite{johnson2019billion}. Our implementation also uses the FAISS library. GPUMAP invokes FAISS through the Python API, missing opportunities for zero-copy exchanges of memory pointers on device \cite{raschka2020machine} that our implementation leverages. 

Manifold learning algorithms typically use the trustworthiness~\cite{venna2006local} score to evaluate a trained model's preservation of local neighborhood structure. The trustworthiness score penalizes divergences in the nearest neighbors between the algorithm's input and output, ranking the similarities of the neighborhoods. Scikit-learn~\cite{pedregosa2011scikit} provides an implementation of  trustworthiness, but the computational costs and memory footprint associated with computing the entire $n^2$ pairwise distance matrix makes it prohibitively slow to evaluate datasets greater than a couple thousand samples.
To the best of our knowledge, there are no existing ports of the trustworthiness score to the GPU. We fill this gap with our new batchable implementation, which we demonstrate can scale well over 100s of thousands of samples on a single GPU with reasonable performance.

\section{Uniform Manifold Approximation and Projection}
\label{sec:umap}

Like many manifold learning algorithms, the UMAP algorithm can be decomposed into three major stages, which we briefly describe in this section. For explanations, derivations, and further details, we refer the reader to the official UMAP paper~\cite{mcinnes2018umap}.

In the first stage, a k-nearest neighbors ($k$-NN) graph is constructed using a distance metric, $d(x,y)$. The second stage weights the closest neighbors around each vertex in the nearest neighbors graph, converting them to fuzzy sets and combining them into a fuzzy union. The fuzzy set membership function learns a locally adaptive exponential kernel that smoothes the distances in each local neighborhood of the $k$-NN graph by finding a smoothing normalizer $\sigma_i$ such that Equation \ref{eq:membership} is satisfied. $\rho$ in this equation contains the distances to the closest non-zero neighbor around each vertex. The triangular conorm \cite{klement1997triangular,dubois1982class} in Equation \ref{eq:conorm} combines the matrix of individual fuzzy sets, $A$, into a fuzzy union by symmetrizing the graph and adding the element-wise (Hadamard) product.
\begin{equation} 
\label{eq:membership}
\sum_{j=i}^k{\exp(-\max(0, d(x_i, x_{i_j})-\rho_i)\sigma_i^{-1})} = \log_2(k)
\end{equation} 
\begin{equation}
\label{eq:conorm}
B = (A + A^T) + (A \circ A^T)
\end{equation}


In the third and final stage, the embeddings are laid out in the topological space using stochastic gradient descent. An initial layout is performed either by sampling embeddings from a uniform distribution or computing a spectral embedding over the fuzzy union. The cross-entropy objective function $-\sum_{a,b\in B}{\left( \log(\Phi(a,b)) + \sum_{c \in B}^{m}{\log(1-\Phi(a,c))}\right)}$ 
is minimized over the edges of the fuzzy union, $B$, from Equation \ref{eq:conorm}. This is done with negative sampling  where $m$ in is the number of negative samples per edge.
$\Phi$ (Equation \ref{eq:phi}) is the current membership strength in the newly embedded space and $\mli{min\_dist}$ controls the minimum separation distance between points in the embedded space. We use the approximate form of $\Phi$ in this paper for simplicitly. The $\log(\Phi)$ term in the objective is computed using the source and destination vertices on each edge and $\log(1-\Phi)$ is computed using the source vertex with negative sampling.

\begin{equation}
\label{eq:phi}
\begin{scriptsize}
\Phi(x,y) \approx \left\{
        \begin{array}{ll}
            1 & \left\|x-y\right\|_2 \le \mli{min\_dist} \\
            \exp(-\left\|x-y\right\|_2-\mli{min\_dist}) & \mli{otherwise}
        \end{array}
    \right\}
\end{scriptsize}
\end{equation} 

When training labels are provided, an additional step in the neighborhood weighting stage adjusts the membership strengths of the fuzzy sets based on their labels. In addition to its learned parameters, the trained UMAP model keeps a reference to the $k$-NN index computed on the training data. This is used to create a mapping of the trained embeddings to a new set of vertices during inference.

\section{GPU-Accelerating UMAP} \label{sec:gpu_acceleration}

Our implementation is primarily written in C++, which is wrapped in a Python API through the Cython library. Data can be passed into our Python API using common formats like Numpy~\cite{Harris2020} or Pandas~\cite{mckinney2011pandas}, as well as GPU array libraries such as CuPy~\cite{okuta2017cupy}, Numba~\cite{lam2015numba}, or RAPIDS cuDF~\cite{raschka2020machine}. When necessary, the data is automatically copied onto the device (e.g., when a Numpy array is passed in). Columnar memory layouts, such as those used in Apache Arrow~\cite{nuggets2017apache}, tend to exploit the optimized memory access patterns on GPUs, such as coalesced accesses~\cite{davidson1994memory}. Like UMAP-learn, our Python API maintains compatibility with the Scikit-learn~\cite{pedregosa2011scikit} API. We used the RAPIDS memory manager (RMM) to create a single memory pool for each process to avoid device synchronization from allocations and deallocations of temporary device memory. When available, we made use of existing libraries with optimized CUDA primitives, such as Thrust~\cite{bell2012thrust}, cuSparse~\cite{naumov2010cusparse,li2015comparison}, cuGraph, and cuML~\cite{ocsa2019sql,raschka2020machine}.

Our implementation begins with a straightforward port of UMAP-learn to CUDA/C++, diverging from the design of UMAP-learn only where we found a significant benefit to performance or the memory footprint. The prior GPUMAP implementation attempted a direct conversion of the code design, using Numba CUDA-JIT~\cite{oden2020lessons} functions and CuPy, without any significant diversions. While hypothetically easier to maintain, we will show this produces results that do not always match the original implementation, and does not deliver meaningful speedups. In each section below, we will detail some of the major design choices that make our GPU implementation faster and more faithful to the original results. 

Copying data between host memory and a GPU device comes at a high cost and can quickly become a bottleneck. Transferring 1MB of data can take several hundreds of microseconds. Even when memory is copied asynchronously between device and host, the underlying CUDA stream needs to be synchronized before the data can be used on the host, further increasing this latency. We reduce the need to transfer between host and device as much as possible, even if that means running code on the GPU that has little to no speedup over the CPU \cite{harris_2012}. Standards like the \textit{\_\_cuda\_array\_interface\_\_}~\cite{raschka2020machine} and \textit{dlpack}~\cite{dmlcdlpa99:online} enable Python libraries to share CUDA memory pointers directly, without the need for copies \cite{raschka2020machine}. This further reduces the need for transfers to host, and like the standard \textit{\_\_array\_interface\_\_}~\cite{Harris2020}, enables implicit conversion between types even as data is passed between different libraries.

\subsection{GPU Architecture}

The NVIDIA General-purpose GPU computing architecture \cite{owens2008gpu,luebke2008cuda} enables parallelism through the single-instruction multiple data design paradigm \cite{raschka2020machine}. A single GPU device contains several banks of global memory that are accessible from a grid containing thousands of instruction processing cores called symmetric-multiprocessors, or SMs, which execute blocks of threads, called thread-blocks, in parallel.  

Each SM has its own faster but smaller bank of memory, called shared memory, which is partitioned across the thread-blocks it executes and enables a series of concurrently executing threads within each thread-block to share data. Each SM executes groupings of 32 threads, known as warps, on the physical hardware. A scheduler logically groups warps into blocks based on the configured block size of a CUDA kernel. Each thread has a series of registers available for storing local variables, which can also be shared across the threads within each warp. 

SMs limit the number of warps and blocks each can execute concurrently. The total number of registers and amount of shared memory available is also limited by each SM, the amount provided to each thread-block depending largely on the block size, or number of threads, configured for each. The amount of shared memory and number of registers used further impacts the number of warps and blocks that can be scheduled concurrently on each SM. These details provide a set of knobs that designers of CUDA kernels can use to optimize their use of the resources available.

GPUs provide the most performance gains when memory access patterns are able to take advantage of features like collective warp-level operations on registers, shared memory, uniform conditional branching, and coalesced memory accesses. Though core-for-core typically not as fast as a CPU, parallelizing operations over GPU threads can still provide significant performance gains even when memory access patterns and the uniformity of computations across threads are not efficient \cite{harris_2012}.

\subsection{Constructing the World $k$-NN Graph}
\label{sec:construct_knn_graph}

The UMAP-learn library utilizes nearest neighbors descent \cite{dong2011efficient} for construction of an approximate nearest neighbors graph, however no known GPU-accelerated versions of this algorithm exist at the time of writing. Tree-based approximate variants, such as the algorithms available in Scikit-learn, also don't have a straightforward port to the GPU \cite{wieschollek2016efficient}. This is a direct result of the iterative nature of traversal, as well as the storage and representation requirements for the trees after they are constructed.

Our implementation of UMAP makes use of the FAISS library \cite{johnson2019billion} for fast nearest neighbors search on GPUs. Other GPU-acceleratd manifold implementations have used this same approach (e.g., t-SNE \cite{chan2018t}). 
FAISS provides both exact and approximate methods to nearest neighbors search, the former being used by default in our implementation. We use the exact search provided by FAISS since it is performant and doesn't require underlying device memory be copied during the hand-off.

For smaller datasets of a few hundred thousand samples and a few hundred features, we found the quadratic scale of the exact $k$-NN graph computation to comprise ~26\% of the total time UMAP spends in compute, making it the second largest performance bottleneck next to the optimization of the embeddings. However, as we demonstrate in Figure \ref{fig:google_scale}, the $k$-NN graph can quickly become the largest bottleneck as the number of samples and features increase to larger sizes, while the optimization stage consistently maintains high performance. This is not a surprising find, since the brute force approach requires exhaustive distances to be computed along with a heap, which is required to maintain the sorted order of closest neighbors. An additional cause of significant performance degradation during this stage is FAISS' incompatibility with outside memory managers, causing unavoidable and expensive synchronous device memory allocations and deallocations for temporary scratch space. We extended the API to accept a $k$-NN graph that has already been computed. This modification provides a strategy to avoid these expensive synchronizations in exploratory environments where a model might be trained several times on the same data.

\subsection{Handling Sparse Data} \label{sec:handling_sparse}

Once computed, many operations are performed over the sparse $k$-NN graph. This is common in many modern manifold learning approaches as well as network analysis problems, where these performance optimizations may be reused. The edges of the $k$-NN graph are unidirectional and the index and distance arrays represent the column and data arrays of a sparse format. The fixed degree makes the row array implicit and allows the use of dense matrix operations until the construction of the fuzzy union, where the degree is no longer fixed.

\begin{figure}[!htbp]
    \centering
        \includegraphics[width=\columnwidth]{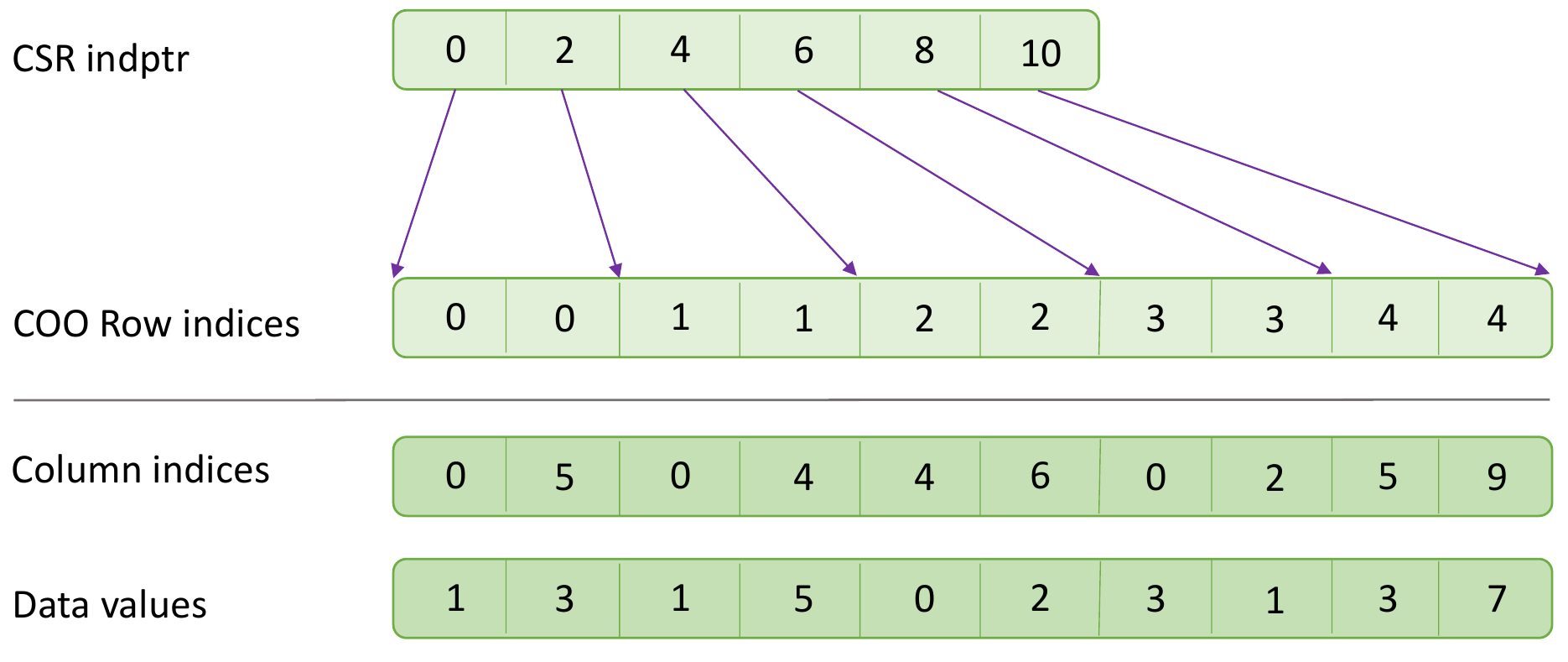}
    \caption{Example of our CSR index being used to index into a sorted COO index.}
    \label{fig:sparse_formats}
\end{figure}

We use the COOrdinate (COO), or edge list format for efficient out-of-order parallel consruction and subsequent element-wise operations. We have found sorting an out-of-order COO can take up to 3\% of the total time spent in compute. When it is efficient to do so, we sort the COO arrays by row and create a Compressed Sparse Row (CSR) index into the column and data arrays, enabling both efficient row- and element-wise parallelism so long as the sorted order is maintained. See Figure \ref{fig:sparse_formats} for a diagram. 

While our implementation makes use of libraries like cuSparse and cuGraph for common operations on sparse data, we built several reusable primitives for operations such as sparse $L_1$ and $L_\infty$ normalization, removal of nonzeros, and the symmetrization required for computing the triangular conorm, where existing implementations of these operations were not available. Aside from custom kernels that don't have much potential for reuse outside of UMAP, such as those described in the following three sections, reusable primitives comprise a large portion of the algorithm.

\subsection{Neighborhood Weighting}

The neighborhood weighting step begins with constructing the $\rho$ and $\sigma$ arrays, with one element for each vertex of the $k$-NN graph. $\rho$ contains the distance to the closest neighbor of each vertex and $\sigma$ contains the smoothing approximator to the fuzzy set membership function for the local neighborhoods of each source vertex in the $k$-NN graph. The operations for computing these two arrays are fused into a single kernel, which maps each source vertex of the $k$-NN graph in CSR format to a separate CUDA thread. The computations in this kernel are largely similar to corresponding Python code in the reference implementation and comprise less than 0.1\% of the total time spent in compute.

The $k$-NN distances are weighted by applying the fuzzy set membership function from the previous step to the COO matrix containing the edges of each source vertex in the $k$-NN graph. Since this computation requires no dependencies between the edges in the neighborhood graph, the CUDA kernel maps each neighbor to their own thread individually.

As described in Section \ref{sec:umap}, the final step of the neighborhood weighting stage combines all the fuzzy sets, using the triangular conorm to build a fuzzy union. We implemented this step by fusing both symmetrization sum and product steps together into a single kernel, using the CSR indptr we introduced in Section \ref{sec:handling_sparse} as a Compressed Sparse Column (CSC) indptr to look up the the transposed value and apply the triangular conorm to each element in parallel. This step comprises less than 0.2\% of the total time spent in compute.

Larger kernels composed of smaller fused operations, such as computing the mean, min, and iterating for the adaptive smoothing parameters, allowed us to make use of registers where the alternative required intermediate and more expensive storage. We found a 12-15$\times$ speedup for the adaptive smoothing operations when compared to separate kernels that require intermediate results to be stored in global memory and accessed without memory coalescing. The end-to-end neighborhood weighting stage exploits parallelism at the expense of potential thread divergence from non-uniform conditional branching, and help the kernels to stay compute-bound.

\subsection{Embedding Updates} \label{sec:embedding_updates}

The first step of the embeddings optimization stage initializes the array of output embeddings. We provide both random and spectral intialization strategies. While the reference implementation uses a spectral embedding of the fuzzy union through the nearest-neighbors variant of the Laplacian eigenmaps \cite{belkin2002laplacian} algorithm, we use the spectral clustering implementation from cuGraph~\cite{fender2017parallel}, setting the number of clusters to 1 and removing the lowest eigenpairs. We have found spectral clustering to be sufficient for maintaining comparable trustworthiness in our experiments while comprising less than 0.1\% of the total time spent in compute.

The optimization step performs stochastic gradient descent over the edges of the fuzzy union, minimizing the cross entropy described in Section \ref{sec:umap}. The gradient computation and update operations have been fused into a single kernel and parallelized so that each thread processes one edge of the fuzzy union. The CUDA kernel is scheduled iteratively for $\mli{n\_epochs}$ to compute and apply the gradient updates to the embeddings in each epoch. The dependencies between the vertices in the updating of the gradients makes this step non-trivial to parallelize efficiently, which decreases potential for coalesced memory access and creates the need for atomic operations when applying gradient updates. As a result, we have seen this kernel take up to 30\% of the total time spent in compute for datasets of a few hundred thousand samples with a few hundred features. When the $k$-NN graph is pre-computed, this step can comprise up to 50\% of the remaining time spent in compute. The dependencies between vertices also create challenges to reproduciblity, which we describe in Section \ref{sec:reproducibility}.

Both the source and destination vertices are updated for each edge during training. Since the trained embeddings should remain unchanged, only the destination vertex is updated during inference. In addition, both training and inference require the source vertex be updated for some number of randomly sampled vertices. Each source vertex will perform $\mli{n\_components} * (\mli{n\_negative\_samples} + 1)$ atomic writes in each thread plus an additional write for the destination vertex during training.

When $\mli{n\_components}$ is small enough, such as a few hundred, we use shared memory to create a small local cache per compute thread, accumulating the updates for each source vertex from multiple negative samples before writing the results atomically to global memory. When shared memory can be used, this reduces atomic updates per thread by a factor of $\mli{n\_components} * \mli{n\_negative\_samples}$. We have measured performance gains of 10\% for this stage when $\mli{n\_components}=2$ to $56\%$ when $\mli{n\_components}=16$ and expect the performance benefits to continue increasing in proportion to $\mli{n\_components}$. For these cases where $\mli{n\_components}$ is very small, such as $\mli{n\_components}=2$, these updates can be accumulated right in the registers, providing a speedup of 49\% for this stage. We suspect these strategies, and any future optimizations, will be useful broadly given the many algorithms (e.g., word2vec~\cite{Mikolov2013a}) that make use of negative sampling.

\subsection{Reproducibility} \label{sec:reproducibility}

Following the original implementation of UMAP, the user can provide a seed to control the random intialization of weights to increase the reproducibility. This does not eliminate all inconsistency when working with parallel updates made from multiple threads. When using a limited number of CPU cores ($\leq 40$ in most circumstances), this effect is minimal. However, with a GPU that has thousands of parallel threads, even subtle timing differences between the thread-blocks can have a large impact on the consistency of results. In addition, large numbers of updates can become queued waiting to be performed atomically. A similar issue is observed with the Hogwild algorithm even when atomic updates are used~\cite{zhang2016hogwild++,NIPS2011_4390,Hsieh:2015:PPA:3045118.3045370,raff_saus,scaling-up-stochastic-dual-coordinate-ascent,Chin:2015:FPS:2745393.2668133}, but at a larger scale. This problem is further exacerbated by small divergences in the processing of instructions that results from non-uniform conditional branching across threads.

Our use of a local cache to accumulate updates as described in Section \ref{sec:embedding_updates} alleviates this by decreasing the number of global atomic writes, helping to reduce the potential for thread divergence and resulting in higher quality solutions. While this minimizes the writes significantly, we still found the potential for inconsistencies to increase in proportion to the number of vertices in the dataset, the number of edges in the fuzzy union, and the number of components being trained.

The results are made fully repeatable with exact precision by optionally using a 64-bit float array to accumulate the updates to the embeddings and applying the updates at the end of each epoch. The additional precision avoids the numerical instabilities created by repeatedly summing small values in a finite range while the single application of the updates removes the potential for race conditions between reads and writes \cite{villa2009effects}. We found the performance impact to increase with the number of components, from an end-to-end slowdown of $11\times$ with $\mli{n\_components}=2$ to $20\times$ with $\mli{n\_components}=16$ on a Volta GV100 GPU.

\subsection{Distributed Inference}
\label{sec:distributed_umap}

Because of its ability to embed out-of-sample data points \cite{NIPS2003_2461}, we scaled the UMAP algorithm to support datasets larger than a single GPU by training a model on a random sample of the training dataset, sending the trained embeddings to a set of workers, each mapped to its own GPU, and performing inference in parallel on the remaining data samples. Our implementation minimizes the use of host memory during communication by using CUDA IPC \cite{potluri2012optimizing} to support fast communication over NVLink~\cite{li2019evaluating} internal to a physical machine and GPUDirect RDMA \cite{venkatesh2014high} to communicate across machine boundaries. We use the Dask library, which has been GPU-accelerated \cite{raschka2020machine} and optimized with the Unified-Communications-X library (UCX) \cite{shamis2015ucx} to support CUDA IPC and GPUDirect transports automatically, without the need to invoke the aforementioned transports directly.

We have found our distributed implementation to scale linearly with the number of GPUs and find it can acceptably preserve structure for a small single-cell RNA dataset containing only 23 thousand cells when trained on as little as 3\% of the data with less than a 1\% drop in trustworthiness. Further, we find only a 0.05\% drop in trustworthiness when we embed the remaining 97\% of the dataset over 16 separate workers.

\begin{table*}[!h]
\caption{Each result shows mean $\pm$ variance, followed by max trustworhiness score, of each implementation of UMAP for the unsupervised case with default parameters. Fastest result in \textbf{bold}.} 
\label{tbl:results_unsupervised}
\centering
\adjustbox{max width=\textwidth}{%
\begin{tabular}{@{}lcccccc@{}}
\toprule
\multicolumn{1}{c}{} & \multicolumn{2}{c}{UMAP-Learn} & \multicolumn{2}{c}{GPUMAP}   & \multicolumn{2}{c}{\cuML}             \\ \cmidrule(lr){2-3} \cmidrule(lr){4-5} \cmidrule(l){6-7} 
Dataset              & $\mu \pm \sigma^2$    & Trust\%  & $\mu \pm \sigma^2$   & Trust\% & $\mu \pm \sigma^2$           & Trust\% \\ \midrule
digits               & $6.328 \pm 2.897$   & 98.79    & $2.483 \pm 1.058$  & 95.58   & \textbf{0.3583$\pm$0.0111} & 98.77   \\
fashion mnist        & $45.87 \pm 10.23$   & 97.81    & $4.158 \pm 1.800$  & 97.50   & \textbf{0.455$\pm$0.006}   & 97.73   \\
mnist                & $52.575 \pm 1.1677$   & 95.94    & $10.6071 \pm 0.45444$  & 94.43   & \textbf{0.70781$\pm$0.0088}   & 95.74   \\
cifar100             & $105.85 \pm 2.482$  & 84.72    & $6.186 \pm 1.770$  & 84.01   & \textbf{1.009$\pm$0.0188}  & 83.42   \\
coil20               & $11.210 \pm 2.571$  & 99.36    & $2.582 \pm 0.0050$ & 95.67   & \textbf{0.757$\pm$0.5752}  & 99.28   \\
shuttle              & $38.88 \pm 8.039$   & 100.0    & $9.064 \pm 3.431$  & 97.78   & \textbf{0.5825$\pm$0.0252} & 100.0   \\
scRNA            & $223.9 \pm 9.071$   & 62.38    & $10.89 \pm 1.604$  & 94.35   & \textbf{4.103$\pm$0.0601}  & 97.81   \\ 
\bottomrule
\end{tabular}
}
\end{table*}

\begin{table*}[!h]
\caption{Each result shows mean $\pm$ variance, followed by max trustworhiness score, of each implementation of UMAP for the supervised case with default parameters. Fastest result in \textbf{bold}.}
\label{tbl:results_supervised}
\centering
\adjustbox{max width=\textwidth}{%
\begin{tabular}{@{}lcccccc@{}}
\toprule
\multicolumn{1}{c}{} & \multicolumn{2}{c}{UMAP-Learn} & \multicolumn{2}{c}{GPUMAP}   & \multicolumn{2}{c}{\cuML}             \\ \cmidrule(lr){2-3} \cmidrule(lr){4-5} \cmidrule(l){6-7} 
Dataset              & $\mu \pm \sigma^2$    & Trust\%  & $\mu \pm \sigma^2$   & Trust\% & $\mu \pm \sigma^2$           & Trust\% \\ \midrule
digits               & $6.756 \pm 0.1109$  & 98.76    & $2.553 \pm 1.095$  & 95.55   & \textbf{0.4063$\pm$0.0135} & 98.80   \\
fashion mnist        & $53.09 \pm 6.183$   & 97.81    & $6.477 \pm 0.0632$ & 96.97   & \textbf{1.0370$\pm$0.0002} & 97.76   \\
mnist                & $89.1877 \pm 6.4658$  & 95.85    & $23.905 \pm 7.057$ & 94.69   & \textbf{0.9175$\pm$0.00297} & 95.74   \\
cifar100             & $98.42 \pm 2.273$   & 84.91    & $5.954 \pm 0.0236$ & 83.01   & \textbf{1.0816$\pm$0.0003} & 83.82   \\
coil20               & $12.34 \pm 0.0217$  & 98.68    & $8.210 \pm 0.0275$ & 93.33   & \textbf{0.3695$\pm$0.0066} & 98.70   \\
shuttle              & $50.17 \pm 17.55$   & 100.0    & $17.15 \pm 23.92$  & 96.67   & \textbf{0.5560$\pm$0.0111} & 100.0   \\ \bottomrule
\end{tabular}
}
\end{table*}

\section{Experiments}
\label{sec:experiments}

\begin{table}[!h]
\caption{Datasets used in experiments} \label{tbl:results}
\centering
\begin{tabular}{@{}l@{}rrr@{}}
\toprule
Dataset                    & Rows & Cols & Classes \\ \midrule
Digits~\cite{garris1994nist}                 & 1797 & 64 & 10  \\ 
Shuttle~\cite{ucishuttle2020}                    & 58k  & 9 & 7  \\
Fashion MNIST\\~\cite{xiao2017fashion}              & 60k  & 784 & 10   \\
MNIST~\cite{deng2012mnist}                      & 60k & 784 & 10 \\
CIFAR-100~\cite{krizhevsky2009learning}                  & 60k  & 1024 & 20  \\
COIL-20~\cite{nene1996columbia} & 1440 & 16384 & 20 \\
scRNA~\cite{travaglini2019molecular}                      & 64.5k   & 5k & N/A \\
GoogleNews Word2vec\\~\cite{mikolov2013efficient}        & 3M & 300 & N/A \\
\bottomrule
\end{tabular}
\end{table}

We compare the execution time and correctness of GPUMAP and our implementation against the multi-core implementation of UMAP-learn on CPU. The datasets are summarized in Table \ref{tbl:results}, showing the number of rows, columns, and classes. We evaluated the execution times of unsupervised training on each dataset for all three implementations and recorded the resulting times, in seconds. Where classes were provided, we also evaluated the supervised training mode. All experiments were conducted on a single DGX1 containing 8 Nvidia GV100 GPUs with Dual Intel Xeon 20-core CPUs. UMAP-learn was configured to take advantage of all the available threads on the machine. 

\begin{table}[!h]
\caption{Execution times for computing the Trustworthiness score with UMAP's default of $\mli{n\_neighbors}=15$.
The first column shows the number of samples used, and the right two columns present run time in seconds. The number of features was fixed to $1024$.
Best results are in \textbf{bold}.  
Done using isotropic blobs
} 
\label{tbl:trust_speedup} 
\centering
\adjustbox{max width=\columnwidth}{%
\begin{tabular}{rrr}
\toprule
Samples                    & Scikit-learn & \cuML   \\ 
\midrule
2k              & 0.33       & \textbf{0.13}  \\
5k                   & 2.06    & \textbf{0.18}  \\
10k                   & 8.64    & \textbf{0.24}   \\
20k                   & 35.76    & \textbf{0.54}   \\
50k                   & 303.08    & \textbf{2.07}   \\
100k                   & FAIL    & \textbf{5.74}   \\
1M                   & FAIL    & \textbf{446.26}   \\
\bottomrule
\end{tabular}
}
\end{table}

We use trustworthiness to rank the degree to which local neighborhood structure is preserved between input and embedded spaces. Scikit-learn provides an implementation of this score, but the execution time and memory requirement of computing the pairwise distance matrix make it prohitive on some of the datasets used in this paper. We implemented a batched GPU-accelerated version of trustworthiness that provides reasonably low execution times for datasets up to 1M samples. Table \ref{tbl:trust_speedup} contains the execution times of computing the trustworthiness score on various different numbers of samples in both Scikit-learn and \cuML. This contribution was necessary to perform our evaluations and was used to evaluate the correctness of each implementation. Because users often select the result with the highest trustworthness score, we report the max score in results.

We begin by demonstrating the speedups obtained by our new \cuML{}  implementation of UMAP in the standard unsupervised scenario. The timing results with standard deviation from 4 runs can be found in Table \ref{tbl:results_unsupervised}, with the trustworthiness score on the right. \cuML{} dominates all other implementations in speed, with $17\times$ speeupds compared to UMAP-learn on the smallest datasets, and increasing to up to $104.9\times$ on moderate scale datasets like MNIST. Similar results can be seen in the supervised case in Table \ref{tbl:results_supervised}.
\cuML{} is also $2.65-15.6\times$ faster than the prior GPUMAP, with an average $7.29\times$ advantage. This is biased towards GPUMAP's favor by the fact that it has regressions in solution quality, as measured by trustworthiness, on 6/7 datasets. 

The trustworthness and speedups show the value of our contributions in Section \ref{sec:gpu_acceleration}, and we, in addition, note that the CPU-based UMAP-learn has its own regression on the scRNA dataset. This is a known issue caused by the lack of synchronized updates in its implementation\footnote{see \url{https://umap-learn.readthedocs.io/en/latest/reproducibility.html}}, following the hog-wild style update of parameters \cite{NIPS2011_4390}. This dataset's large number of features and datapoints combine to create race conditions that are too significant for correct results. This shows the importance of our register accumulation strategy introduced in Section \ref{sec:reproducibility}, allowing us to obtain better quality results in these extreme cases.

\begin{figure}[!htbp]
    \centering
        \includegraphics[width=\columnwidth]{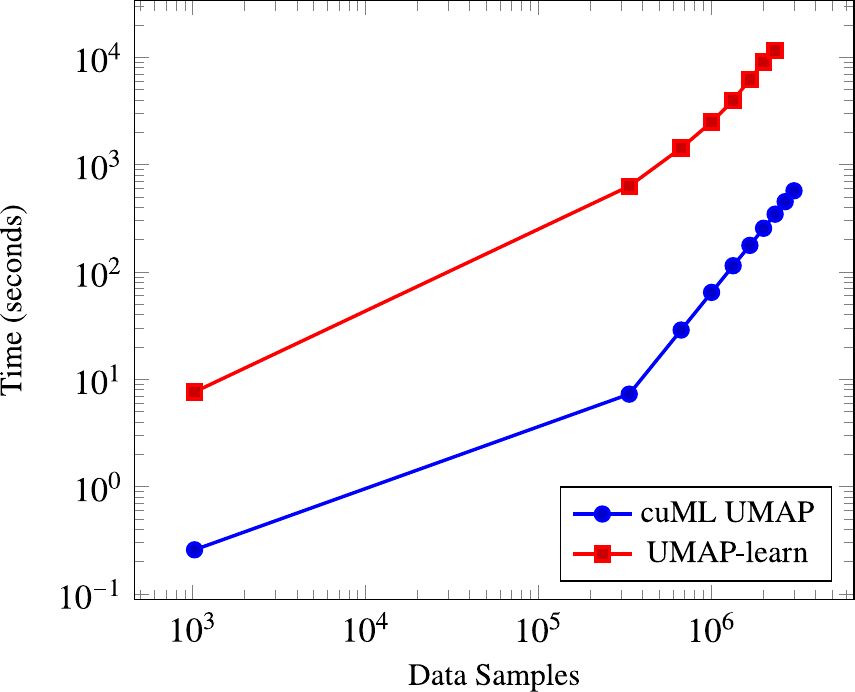}
    \caption{Google-News Results showing runtime (y-axis) as more of the dataset is sampled (x-axis).}
    \label{fig:google_scale}
\end{figure}

    
    
    



The original GPUMAP implementation has, at times, had runtime failures where no results are produced, and significant time was spent attempting to re-compile/fix these issues without success. This prevented its use on our largest dataset, Google-News word2vec embeddings. We use the Google-News corpus in particular as a large-scale experiment to show the value of our results, compared to UMAP-learn up to a time limit of 3 hours. The runtime comparing many cores with UMAP-learn to our \cuML{} on a single GPU is shown in Figure \ref{fig:google_scale}. We can clearly see that \cuML{} continues to dominate runtime with no loss in quality, obtaining $\geq 30\times$ speeupds across $n=1,024$ samples all the way up to the full 3M samples while UMAP-learn reached its time limit after 2.3M samples.

Our \cuML's 9.5 minutes to process all 3 million datapoints of Google-News is already a significant advantage in runtime. In addition, we note that all non $k$-NN work of the UMAP algorithm took only \textit{9.3 seconds} of that total time. This is important for the interactive and hyper-parameter tuning scenarios. Our implementation allows computing the $k$-NN once, and then other hyperparameter settings can be adjusted with results obtained in seconds. Section \ref{sec:distributed_umap} briefly discusses our distributed UMAP inference algorithm, with preliminary results demonstrating an ability to embed 10M points (including $k$-NN) across 8 GPUs in just under 5 seconds with only a marginal impact to trustworthiness. These results can speedup tuning and visualization by orders of magnitude, and is enabled by the optimizations we have contributed.

We also note that the consistent speedups on both small and large datasets is important for showing the isoefficiency~\cite{Grama:1993:IMS:613769.613817} of our method. A poorly implemented parallel method may exhibit speedups if sufficently large amounts of data/work are fed to counter-balance the overheads of communication primitives used. Since we obtain speedups on small datasets, our results are demonstrating good isoefficiency and users should be able to regularly obtain a better runtime by using our method --- rather than having to judge if the dataset is "big enough" to make our implementation worth while. These speedups on small datasets are also important for practioners where the difference between 1 minute and 1 second are noticible and additionally enables the computer-human interaction usecases we hope to enable with this work.

\section{Conclusion} \label{sec:conclusion}

The UMAP algorithm is becoming a widely popular tool, which increases the demand and utility of a faster implementation. We have detailed a number of techniques that are easy to apply in code, and allow us to obtain a solution that is faster and more accurate, even at times compared to the original CPU-based implementation. This obtains up to 100$\times$ speedups, and by eliminating all non $k$-NN calculations to $\leq 2$\% of runtime, we enable interactive exploration and parameter tuning use-cases that were previously untenable.

\section*{Acknowledgement}

We extend our sincerest gratitude to all of those who helped enable our research, especially Philip Hynsu Cho and Dante Gama Dessavre from the RAPIDS cuML team as well as Brad Rees, Alex Fender, and Joe Eaton from the RAPIDS cuGraph team. We would also like to thank the Clara Genomics team at Nvidia, especially Avantika Lal, Johnny Israeli, Raghav Mani, and Neha Tadimeti. In addition, we thank Jeff Johnson \& Matthijs Douze of the FAISS project for their continued support and Dmitri Kobak, whose single-cell RNA preprocessing scripts were helpful in our evaluations. Finally, we owe our deepest thanks to Leland McInnes \& John Healy, because this research would not exist without their contributions.

{\fontsize{9pt}{10pt} \selectfont
\bibliography{sample-base}}

\begin{thebibliography}{75}
\providecommand{\natexlab}[1]{#1}
\providecommand{\url}[1]{\texttt{#1}}
\providecommand{\urlprefix}{URL }
\expandafter\ifx\csname urlstyle\endcsname\relax
  \providecommand{\doi}[1]{doi:\discretionary{}{}{}#1}\else
  \providecommand{\doi}{doi:\discretionary{}{}{}\begingroup
  \urlstyle{rm}\Url}\fi

\bibitem[{Akiba et~al.(2019)Akiba, Sano, Yanase, Ohta, and
  Koyama}]{Akiba:2019:ONH:3292500.3330701}
Akiba, T.; Sano, S.; Yanase, T.; Ohta, T.; and Koyama, M. 2019.
\newblock {Optuna: A Next-generation Hyperparameter Optimization Framework}.
\newblock In \emph{KDD}, 2623--2631. ACM.

\bibitem[{Andor et~al.(2018)Andor, Lau, Catalanotti, Kumar, Sathe, Belhocine,
  Wheeler, Price, Song, Stafford et~al.}]{andor2018joint}
Andor, N.; Lau, B.~T.; Catalanotti, C.; Kumar, V.; Sathe, A.; Belhocine, K.;
  Wheeler, T.~D.; Price, A.~D.; Song, M.; Stafford, D.; et~al. 2018.
\newblock Joint single cell DNA-Seq and RNA-Seq of gastric cancer reveals
  subclonal signatures of genomic instability and gene expression.
\newblock \emph{bioRxiv} 445932.

\bibitem[{Becht et~al.(2018)Becht, Dutertre, Kwok, Ng, Ginhoux, and
  Newell}]{becht2018evaluation}
Becht, E.; Dutertre, C.-A.; Kwok, I.~W.; Ng, L.~G.; Ginhoux, F.; and Newell,
  E.~W. 2018.
\newblock Evaluation of UMAP as an alternative to t-SNE for single-cell data.
\newblock \emph{BioRxiv} 298430.

\bibitem[{Belkin and Niyogi(2002)}]{belkin2002laplacian}
Belkin, M.; and Niyogi, P. 2002.
\newblock Laplacian eigenmaps and spectral techniques for embedding and
  clustering.
\newblock In \emph{NeurIPS}, 585--591.

\bibitem[{Bell and Hoberock(2012)}]{bell2012thrust}
Bell, N.; and Hoberock, J. 2012.
\newblock Thrust: A productivity-oriented library for CUDA.
\newblock In \emph{GPU computing gems Jade edition}, 359--371. Elsevier.

\bibitem[{Bengio et~al.(2004)Bengio, fran\c{c}cois Paiement, Vincent,
  Delalleau, Roux, and Ouimet}]{NIPS2003_2461}
Bengio, Y.; fran\c{c}cois Paiement, J.; Vincent, P.; Delalleau, O.; Roux,
  N.~L.; and Ouimet, M. 2004.
\newblock Out-of-Sample Extensions for LLE, Isomap, MDS, Eigenmaps, and
  Spectral Clustering.
\newblock In \emph{NeurIPS}, 177--184. MIT Press.

\bibitem[{Beygelzimer, Kakade, and Langford(2006)}]{Beygelzimer2006}
Beygelzimer, A.; Kakade, S.; and Langford, J. 2006.
\newblock {Cover trees for nearest neighbor}.
\newblock In \emph{ICML}, 97--104. ACM.

\bibitem[{Carter et~al.(2019)Carter, Armstrong, Schubert, Johnson, and
  Olah}]{carter2019exploring}
Carter, S.; Armstrong, Z.; Schubert, L.; Johnson, I.; and Olah, C. 2019.
\newblock Exploring neural networks with activation atlases.
\newblock \emph{Distill} \doi{10.23915/distill.00015}.

\bibitem[{Chan et~al.(2018)Chan, Rao, Huang, and Canny}]{chan2018t}
Chan, D.~M.; Rao, R.; Huang, F.; and Canny, J.~F. 2018.
\newblock t-SNE-CUDA: GPU-Accelerated t-SNE and its Applications to Modern
  Data.
\newblock In \emph{2018 30th International Symposium on Computer Architecture
  and High Performance Computing (SBAC-PAD)}, 330--338. IEEE.

\bibitem[{Chang and Lin(2011)}]{Chang2011}
Chang, C.-C.; and Lin, C.-J. 2011.
\newblock {LIBSVM: A library for support vector machines}.
\newblock \emph{ACM Transactions on Intelligent Systems and Technology} 2(3).
\newblock \doi{10.1145/1961189.1961199}.

\bibitem[{Chatzimparmpas, Martins, and Kerren(2020)}]{chatzimparmpas2020t}
Chatzimparmpas, A.; Martins, R.~M.; and Kerren, A. 2020.
\newblock t-viSNE: Interactive Assessment and Interpretation of t-SNE
  Projections.
\newblock \emph{arXiv preprint arXiv:2002.06910} .

\bibitem[{Chen(2020)}]{dmlcdlpa99:online}
Chen, T. 2020.
\newblock dmlc/dlpack: RFC for common in-memory tensor structure and operator
  interface for deep learning system.
\newblock \url{https://github.com/dmlc/dlpack}.
\newblock (Accessed on 06/02/2020).

\bibitem[{Chen and Guestrin(2016)}]{xgboost}
Chen, T.; and Guestrin, C. 2016.
\newblock {XGBoost: Reliable Large-scale Tree Boosting System}.
\newblock In \emph{KDD}.

\bibitem[{Chin et~al.(2015)Chin, Zhuang, Juan, and
  Lin}]{Chin:2015:FPS:2745393.2668133}
Chin, W.-S.; Zhuang, Y.; Juan, Y.-C.; and Lin, C.-J. 2015.
\newblock {A Fast Parallel Stochastic Gradient Method for Matrix Factorization
  in Shared Memory Systems}.
\newblock \emph{ACM Transactions on Intelligent Systems and Technology (TIST)}
  6(1): 2:1----2:24.
\newblock \doi{10.1145/2668133}.

\bibitem[{Chollet et~al.(2018)}]{chollet2018keras}
Chollet, F.; et~al. 2018.
\newblock Keras: The python deep learning library.
\newblock \emph{Astrophysics Source Code Library} .

\bibitem[{Clara-Parabricks(2020)}]{clarapar25:online}
Clara-Parabricks, N. 2020.
\newblock Examples of single-cell genomic analysis accelerated with RAPIDS.
\newblock
  \url{https://github.com/clara-parabricks/rapids-single-cell-examples}.
\newblock (Accessed on 06/03/2020).

\bibitem[{Davidson and Jinturkar(1994)}]{davidson1994memory}
Davidson, J.~W.; and Jinturkar, S. 1994.
\newblock Memory access coalescing: a technique for eliminating redundant
  memory accesses.
\newblock \emph{Acm Sigplan Notices} 29(6): 186--195.

\bibitem[{Deng(2012)}]{deng2012mnist}
Deng, L. 2012.
\newblock The mnist database of handwritten digit images for machine learning
  research [best of the web].
\newblock \emph{IEEE Signal Processing Magazine} 29(6): 141--142.

\bibitem[{Dong, Moses, and Li(2011)}]{dong2011efficient}
Dong, W.; Moses, C.; and Li, K. 2011.
\newblock Efficient k-nearest neighbor graph construction for generic
  similarity measures.
\newblock In \emph{WWW}, 577--586.

\bibitem[{Dorogush, Ershov, and Gulin(2017)}]{CatBoost}
Dorogush, A.~V.; Ershov, V.; and Gulin, A. 2017.
\newblock {CatBoost: gradient boosting with categorical features support}.
\newblock In \emph{Workshop on ML Systems at NeurIPS 2017}.

\bibitem[{Dua and Graff(2017)}]{ucishuttle2020}
Dua, D.; and Graff, C. 2017.
\newblock {UCI} Machine Learning Repository.
\newblock \urlprefix\url{http://archive.ics.uci.edu/ml}.
\newblock (Accessed on 03/28/2021).

\bibitem[{Dubois and Prade(1982)}]{dubois1982class}
Dubois, D.; and Prade, H. 1982.
\newblock A class of fuzzy measures based on triangular norms a general
  framework for the combination of uncertain information.
\newblock \emph{International Journal of General Systems} 8(1): 43--61.

\bibitem[{Fan et~al.(2008)Fan, Chang, Hsieh, Wang, and Lin}]{Fan2008}
Fan, R.-E.; Chang, K.-W.; Hsieh, C.-J.; Wang, X.-R.; and Lin, C.-J. 2008.
\newblock {LIBLINEAR: A Library for Large Linear Classification}.
\newblock \emph{JMLR} 9: 1871--1874.

\bibitem[{Fefferman, Mitter, and Narayanan(2016)}]{fefferman2016testing}
Fefferman, C.; Mitter, S.; and Narayanan, H. 2016.
\newblock Testing the manifold hypothesis.
\newblock \emph{Journal of the American Mathematical Society} 29(4): 983--1049.

\bibitem[{Fender(2017)}]{fender2017parallel}
Fender, A. 2017.
\newblock \emph{Parallel solutions for large-scale eigenvalue problems arising
  in graph analytics}.
\newblock Ph.D. thesis, Universit{\'e} Paris-Saclay.

\bibitem[{Garris et~al.(1994)Garris, Blue, Candela et~al.}]{garris1994nist}
Garris, M.~D.; Blue, J.~L.; Candela, G.~T.; et~al. 1994.
\newblock NIST form-based handprint recognition system.
\newblock In \emph{Technical Report NISTIR 5469 and CD-ROM, National Institute
  of Standards and Technology}. Citeseer.

\bibitem[{Grama, Gupta, and Kumar(1993)}]{Grama:1993:IMS:613769.613817}
Grama, A.~Y.; Gupta, A.; and Kumar, V. 1993.
\newblock {Isoefficiency: Measuring the Scalability of Parallel Algorithms and
  Architectures}.
\newblock \emph{IEEE Parallel Distrib. Technol.} 1(3): 12--21.
\newblock \doi{10.1109/88.242438}.

\bibitem[{Harris et~al.(2020)Harris, Millman, van~der Walt, Gommers
  et~al.}]{Harris2020}
Harris, C.~R.; Millman, K.~J.; van~der Walt, S.~J.; Gommers, R.; et~al. 2020.
\newblock {Array programming with NumPy}.
\newblock \emph{Nature} 585(7825): 357--362.
\newblock \doi{10.1038/s41586-020-2649-2}.

\bibitem[{Harris(2012)}]{harris_2012}
Harris, M. 2012.
\newblock NVIDIA Developer Blog.
\newblock
  \urlprefix\url{https://devblogs.nvidia.com/how-optimize-data-transfers-cuda-cc/}.
\newblock (Accessed on 03/28/2021).

\bibitem[{He et~al.(2005)He, Cai, Yan, and Zhang}]{he2005neighborhood}
He, X.; Cai, D.; Yan, S.; and Zhang, H.-J. 2005.
\newblock Neighborhood preserving embedding.
\newblock In \emph{ICCV}, volume~2, 1208--1213.

\bibitem[{Hsieh, Yu, and Dhillon(2015)}]{Hsieh:2015:PPA:3045118.3045370}
Hsieh, C.-J.; Yu, H.-F.; and Dhillon, I.~S. 2015.
\newblock {PASSCoDe: Parallel Asynchronous Stochastic Dual Co-ordinate
  Descent}.
\newblock In \emph{ICML}, 2370--2379.

\bibitem[{Johnson, Douze, and J{\'e}gou(2019)}]{johnson2019billion}
Johnson, J.; Douze, M.; and J{\'e}gou, H. 2019.
\newblock Billion-scale similarity search with GPUs.
\newblock \emph{IEEE Transactions on Big Data} .

\bibitem[{Ke et~al.(2017)Ke, Meng, Finley, Wang, Chen, Ma, Ye, and
  Liu}]{NIPS2017_6907}
Ke, G.; Meng, Q.; Finley, T.; Wang, T.; Chen, W.; Ma, W.; Ye, Q.; and Liu,
  T.-Y. 2017.
\newblock Lightgbm: A highly efficient gradient boosting decision tree.
\newblock volume~30, 3146--3154.

\bibitem[{Klement, Mesiar, and Pap(1997)}]{klement1997triangular}
Klement, E.; Mesiar, R.; and Pap, E. 1997.
\newblock Triangular norms.
\newblock \emph{Tatra Mountains Math. Publ} 13: 169--193.

\bibitem[{Krizhevsky(2009)}]{krizhevsky2009learning}
Krizhevsky, A. 2009.
\newblock \emph{Learning multiple layers of features from tiny images}.
\newblock Master's thesis.

\bibitem[{Lam, Pitrou, and Seibert(2015)}]{lam2015numba}
Lam, S.~K.; Pitrou, A.; and Seibert, S. 2015.
\newblock Numba: A llvm-based python jit compiler.
\newblock In \emph{Proceedings of the Second Workshop on the LLVM Compiler
  Infrastructure in HPC}.

\bibitem[{LeDem(2017)}]{nuggets2017apache}
LeDem, J. 2017.
\newblock Apache Arrow and Apache Parquet: Why We Needed Different Projects for
  Columnar Data, on Disk and In-Memory.
\newblock
  \url{www.kdnuggets.com/2017/02/apache-arrow-parquet-columnar-data.html}.
\newblock Accessed: 03/26/2021.

\bibitem[{Li et~al.(2015)Li, Mazhar, Serban, and Negrut}]{li2015comparison}
Li, A.; Mazhar, H.; Serban, R.; and Negrut, D. 2015.
\newblock Comparison of SPMV performance on matrices with different matrix
  format using CUSP, cuSPARSE and ViennaCL.
\newblock Technical report, Technical Report TR-2015-02.

\bibitem[{Li et~al.(2019)Li, Song, Chen, Li, Liu, Tallent, and
  Barker}]{li2019evaluating}
Li, A.; Song, S.~L.; Chen, J.; Li, J.; Liu, X.; Tallent, N.~R.; and Barker,
  K.~J. 2019.
\newblock Evaluating Modern GPU Interconnect: PCIe, NVLink, NV-SLI, NVSwitch
  and GPUDirect.
\newblock \emph{IEEE Transactions on Parallel and Distributed Systems} 31(1).

\bibitem[{Luebke(2008)}]{luebke2008cuda}
Luebke, D. 2008.
\newblock CUDA: Scalable parallel programming for high-performance scientific
  computing.
\newblock In \emph{2008 5th IEEE international symposium on biomedical imaging:
  from nano to macro}, 836--838. IEEE.

\bibitem[{Maaten and Hinton(2008)}]{maaten2008visualizing}
Maaten, L. v.~d.; and Hinton, G. 2008.
\newblock Visualizing data using t-SNE.
\newblock \emph{JMLR} 9(Nov): 2579--2605.

\bibitem[{McInnes, Healy, and Melville(2018)}]{mcinnes2018umap}
McInnes, L.; Healy, J.; and Melville, J. 2018.
\newblock Umap: Uniform manifold approximation and projection for dimension
  reduction.
\newblock \emph{arXiv preprint arXiv:1802.03426} .

\bibitem[{McKinney et~al.(2011)}]{mckinney2011pandas}
McKinney, W.; et~al. 2011.
\newblock pandas: a foundational Python library for data analysis and
  statistics.
\newblock \emph{Python for High Performance and Scientific Computing} 14(9).

\bibitem[{Mikolov et~al.(2013{\natexlab{a}})Mikolov, Chen, Corrado, and
  Dean}]{mikolov2013efficient}
Mikolov, T.; Chen, K.; Corrado, G.; and Dean, J. 2013{\natexlab{a}}.
\newblock Efficient estimation of word representations in vector space.
\newblock \emph{arXiv preprint arXiv:1301.3781} .

\bibitem[{Mikolov et~al.(2013{\natexlab{b}})Mikolov, Corrado, Chen, and
  Dean}]{Mikolov2013a}
Mikolov, T.; Corrado, G.; Chen, K.; and Dean, J. 2013{\natexlab{b}}.
\newblock {Efficient Estimation of Word Representations in Vector Space}.
\newblock \emph{ICLR} .

\bibitem[{Naumov et~al.(2010)Naumov, Chien, Vandermersch, and
  Kapasi}]{naumov2010cusparse}
Naumov, M.; Chien, L.; Vandermersch, P.; and Kapasi, U. 2010.
\newblock Cusparse library.
\newblock In \emph{GPU Technology Conference}.

\bibitem[{Nene et~al.(1996)Nene, Nayar, Murase et~al.}]{nene1996columbia}
Nene, S.~A.; Nayar, S.~K.; Murase, H.; et~al. 1996.
\newblock Columbia object image library (coil-20) .

\bibitem[{Obermayer et~al.(2020)Obermayer, Holtgrewe, Nieminen, Messerschmidt,
  and Beule}]{obermayer2020scelvis}
Obermayer, B.; Holtgrewe, M.; Nieminen, M.; Messerschmidt, C.; and Beule, D.
  2020.
\newblock SCelVis: exploratory single cell data analysis on the desktop and in
  the cloud.
\newblock \emph{PeerJ} 8: e8607.

\bibitem[{Ocsa(2019)}]{ocsa2019sql}
Ocsa, A. 2019.
\newblock {SQL for GPU Data Frames in RAPIDS Accelerating end-to-end data
  science workflows using GPUs}.
\newblock {LatinX in AI Research at ICML 2019}.
\newblock \urlprefix\url{https://hal.archives-ouvertes.fr/hal-02264776}.
\newblock Poster.

\bibitem[{Oden(2020)}]{oden2020lessons}
Oden, L. 2020.
\newblock Lessons learned from comparing C-CUDA and Python-Numba for
  GPU-Computing.
\newblock In \emph{2020 28th Euromicro International Conference on Parallel,
  Distributed and Network-Based Processing (PDP)}, 216--223. IEEE.

\bibitem[{Okuta et~al.(2017)Okuta, Unno, Nishino, Hido, and
  Loomis}]{okuta2017cupy}
Okuta, R.; Unno, Y.; Nishino, D.; Hido, S.; and Loomis, C. 2017.
\newblock Cupy: A numpy-compatible library for nvidia gpu calculations.
\newblock In \emph{Proceedings of Workshop on Machine Learning Systems
  (LearningSys) in NeurIPS}.

\bibitem[{Ordun, Purushotham, and Raff(2020)}]{ordun2020exploratory}
Ordun, C.; Purushotham, S.; and Raff, E. 2020.
\newblock Exploratory analysis of covid-19 tweets using topic modeling, umap,
  and digraphs.
\newblock \emph{arXiv preprint arXiv:2005.03082} .

\bibitem[{Owens et~al.(2008)Owens, Houston, Luebke, Green, Stone, and
  Phillips}]{owens2008gpu}
Owens, J.~D.; Houston, M.; Luebke, D.; Green, S.; Stone, J.~E.; and Phillips,
  J.~C. 2008.
\newblock GPU computing.
\newblock \emph{Proceedings of the IEEE} 96(5): 879--899.

\bibitem[{Pachev and Lupo(2017)}]{p3732gpu8:online}
Pachev, I.; and Lupo, C. 2017.
\newblock GPUMap: A Transparently GPU-Accelerated Python Map Function.
\newblock \doi{10.1145/3149869.3149875}.

\bibitem[{Paszke et~al.(2019)Paszke, Gross, Massa, Lerer, Bradbury, Chanan,
  Killeen, Lin, Gimelshein, Antiga et~al.}]{paszke2019pytorch}
Paszke, A.; Gross, S.; Massa, F.; Lerer, A.; Bradbury, J.; Chanan, G.; Killeen,
  T.; Lin, Z.; Gimelshein, N.; Antiga, L.; et~al. 2019.
\newblock PyTorch: An imperative style, high-performance deep learning library.
\newblock In \emph{NeurIPS}, 8024--8035.

\bibitem[{Pedregosa et~al.(2011)Pedregosa, Varoquaux, Gramfort, Michel,
  Thirion, Grisel, Blondel, Prettenhofer, Weiss, Dubourg
  et~al.}]{pedregosa2011scikit}
Pedregosa, F.; Varoquaux, G.; Gramfort, A.; Michel, V.; Thirion, B.; Grisel,
  O.; Blondel, M.; Prettenhofer, P.; Weiss, R.; Dubourg, V.; et~al. 2011.
\newblock Scikit-learn: Machine learning in Python.
\newblock \emph{JMLR} 12: 2825--2830.

\bibitem[{Pezzotti et~al.(2018)Pezzotti, Mordvintsev, Hollt, Lelieveldt,
  Eisemann, and Vilanova}]{pezzotti2018linear}
Pezzotti, N.; Mordvintsev, A.; Hollt, T.; Lelieveldt, B.~P.; Eisemann, E.; and
  Vilanova, A. 2018.
\newblock Linear tsne optimization for the web.
\newblock \emph{arXiv preprint arXiv:1805.10817} .

\bibitem[{Potluri et~al.(2012)Potluri, Wang, Bureddy, Singh, Rosales, and
  Panda}]{potluri2012optimizing}
Potluri, S.; Wang, H.; Bureddy, D.; Singh, A.~K.; Rosales, C.; and Panda, D.~K.
  2012.
\newblock Optimizing MPI communication on multi-GPU systems using CUDA
  inter-process communication.
\newblock In \emph{2012 IEEE 26th International Parallel and Distributed
  Processing Symposium Workshops \& PhD Forum}.

\bibitem[{Raff and Sylvester(2018)}]{raff_saus}
Raff, E.; and Sylvester, J. 2018.
\newblock {Linear Models with Many Cores and CPUs: A Stochastic Atomic Update
  Scheme}.
\newblock In \emph{Big Data}, 65--73.
\newblock \doi{10.1109/BigData.2018.8622172}.

\bibitem[{Raschka, Patterson, and Nolet(2020)}]{raschka2020machine}
Raschka, S.; Patterson, J.; and Nolet, C. 2020.
\newblock Machine Learning in Python: Main developments and technology trends
  in data science, machine learning, and artificial intelligence.
\newblock \emph{Information} 11(4): 193.

\bibitem[{Recht et~al.(2011)Recht, Re, Wright, and Niu}]{NIPS2011_4390}
Recht, B.; Re, C.; Wright, S.; and Niu, F. 2011.
\newblock {Hogwild: A Lock-Free Approach to Parallelizing Stochastic Gradient
  Descent}.
\newblock In \emph{NeurIPS}, 693--701.

\bibitem[{Roweis and Saul(2000)}]{roweis2000nonlinear}
Roweis, S.~T.; and Saul, L.~K. 2000.
\newblock Nonlinear dimensionality reduction by locally linear embedding.
\newblock \emph{science} 290(5500).

\bibitem[{Shamis et~al.(2015)Shamis, Venkata, Lopez, Baker, Hernandez, Itigin,
  Dubman, Shainer, Graham, Liss et~al.}]{shamis2015ucx}
Shamis, P.; Venkata, M.~G.; Lopez, M.~G.; Baker, M.~B.; Hernandez, O.; Itigin,
  Y.; Dubman, M.; Shainer, G.; Graham, R.~L.; Liss, L.; et~al. 2015.
\newblock UCX: an open source framework for HPC network APIs and beyond.
\newblock In \emph{IEEE 23rd Annual Symposium on High-Performance
  Interconnects}, 40--43.

\bibitem[{Tasic et~al.(2018)Tasic, Yao, Graybuck, Smith, Nguyen, Bertagnolli,
  Goldy, Garren, Economo, Viswanathan et~al.}]{tasic2018shared}
Tasic, B.; Yao, Z.; Graybuck, L.~T.; Smith, K.~A.; Nguyen, T.~N.; Bertagnolli,
  D.; Goldy, J.; Garren, E.; Economo, M.~N.; Viswanathan, S.; et~al. 2018.
\newblock Shared and distinct transcriptomic cell types across neocortical
  areas.
\newblock \emph{Nature} 563(7729).

\bibitem[{Tenenbaum, De~Silva, and Langford(2000)}]{tenenbaum2000global}
Tenenbaum, J.~B.; De~Silva, V.; and Langford, J.~C. 2000.
\newblock A global geometric framework for nonlinear dimensionality reduction.
\newblock \emph{science} 290(5500): 2319--2323.

\bibitem[{Tran et~al.(2015)Tran, Hosseini, Xiao, Finley, and
  Bilenko}]{scaling-up-stochastic-dual-coordinate-ascent}
Tran, K.; Hosseini, S.; Xiao, L.; Finley, T.; and Bilenko, M. 2015.
\newblock {Scaling Up Stochastic Dual Coordinate Ascent}.
\newblock In \emph{KDD}, 1185--1194.
\newblock \doi{10.1145/2783258.2783412}.

\bibitem[{Travaglini et~al.(2019)Travaglini, Nabhan, Penland, Sinha, Gillich,
  Sit, Chang, Conley, Mori, Seita et~al.}]{travaglini2019molecular}
Travaglini, K.~J.; Nabhan, A.~N.; Penland, L.; Sinha, R.; Gillich, A.; Sit,
  R.~V.; Chang, S.; Conley, S.~D.; Mori, Y.; Seita, J.; et~al. 2019.
\newblock A molecular cell atlas of the human lung from single cell RNA
  sequencing.
\newblock \emph{bioRxiv} 742320.

\bibitem[{Venkatesh et~al.(2014)Venkatesh, Subramoni, Hamidouche, and
  Panda}]{venkatesh2014high}
Venkatesh, A.; Subramoni, H.; Hamidouche, K.; and Panda, D.~K. 2014.
\newblock A high performance broadcast design with hardware multicast and
  GPUDirect RDMA for streaming applications on Infiniband clusters.
\newblock In \emph{21st International Conference on High Performance Computing
  (HiPC)}, 1--10.

\bibitem[{Venna and Kaski(2006)}]{venna2006local}
Venna, J.; and Kaski, S. 2006.
\newblock Local multidimensional scaling.
\newblock \emph{Neural Networks} 19(6-7): 889--899.

\bibitem[{Villa et~al.(2009)Villa, Chavarria-Miranda, Gurumoorthi, M{\'a}rquez,
  and Krishnamoorthy}]{villa2009effects}
Villa, O.; Chavarria-Miranda, D.; Gurumoorthi, V.; M{\'a}rquez, A.; and
  Krishnamoorthy, S. 2009.
\newblock Effects of floating-point non-associativity on numerical computations
  on massively multithreaded systems.
\newblock In \emph{Proceedings of Cray User Group Meeting (CUG)}, 3.

\bibitem[{Wander et~al.(2020)Wander, Vianello, Vollertsen, Westad, Braun, and
  Paul}]{wander2020exploratory}
Wander, L.; Vianello, A.; Vollertsen, J.; Westad, F.; Braun, U.; and Paul, A.
  2020.
\newblock Exploratory analysis of hyperspectral FTIR data obtained from
  environmental microplastics samples.
\newblock \emph{Analytical Methods} 12(6): 781--791.

\bibitem[{Wieschollek et~al.(2016)Wieschollek, Wang, Sorkine-Hornung, and
  Lensch}]{wieschollek2016efficient}
Wieschollek, P.; Wang, O.; Sorkine-Hornung, A.; and Lensch, H. 2016.
\newblock Efficient large-scale approximate nearest neighbor search on the gpu.
\newblock In \emph{CVPR}, 2027--2035.

\bibitem[{Wolf, Angerer, and Theis(2017)}]{wolf2017scanpy}
Wolf, F.~A.; Angerer, P.; and Theis, F.~J. 2017.
\newblock Scanpy for analysis of large-scale single-cell gene expression data.
\newblock \emph{BioRxiv} 174029.

\bibitem[{Xiao, Rasul, and Vollgraf(2017)}]{xiao2017fashion}
Xiao, H.; Rasul, K.; and Vollgraf, R. 2017.
\newblock Fashion-mnist: a novel image dataset for benchmarking machine
  learning algorithms.
\newblock \emph{arXiv preprint arXiv:1708.07747} .

\bibitem[{Zhang, Hsieh, and Akella(2016)}]{zhang2016hogwild++}
Zhang, H.; Hsieh, C.-J.; and Akella, V. 2016.
\newblock {HogWild++: A New Mechanism for Decentralized Asynchronous Stochastic
  Gradient Descent}.
\newblock In \emph{ICDM}.

\end{thebibliography}

\clearpage
\appendix
\onecolumn

\section{API Enhancements}
\label{sec:api_enhancements}
While libraries like Scanpy~\cite{wolf2017scanpy} invoke private Python functions internal to the reference implementation, we maintain compatibility only with Scikit-learn's public estimators \cite{raschka2020machine} interface. For the remainder of this section, we briefly discuss two enhancements to the standard Scikit-learn API that enable interactive data analysis and visualization workflows.

\subsection{Pre-computed $k$-NN Graph}
\label{sec:accept_knn_graph}

As mentioned in Section \ref{sec:construct_knn_graph} and demonstrated in Section \ref{sec:experiments}, the $k$-NN graph construction stage can quickly become the largest bottleneck to the end-to-end algorithm, eclipsing the remaining stages by orders of magnitude. When many UMAP models need to be trained with different parameters, such as in cluster analysis and hyperparameter-tuning environments, it can be very wasteful to recompute the $k$-NN graph when neither the training data, distance metric, nor the $\mli{n\_neighbors}$ parameter have changed. 

We diverge from the reference API but maintain compatibility with the Scikit-learn API by providing an additional $knn\_graph$ parameter to \textit{fit\(\)}, \textit{fit\_transform\(\)}, and \textit{transform\(\)}. This new parameter allows the $k$-NN graph to be computed externally and passed into our API, therefore bypassing the computation altogether. This enhancement also makes our implementation more flexible and extensible, since new $k$-NN libraries can be used, even with distance metrics that are not yet supported.

\subsection{Training Callbacks}

Inspired by deep learning frameworks like Keras~\cite{chollet2018keras}, the UMAP API has been enhanced to accept a custom Python function that will be invoked during each epoch of the embeddings optimization stage. This enhancement provides an opportunity to introspect and potentially manipulate the array of actual embeddings in GPU device memory during training. We have found this to be a useful feature that enable interactive visualization tools to provide visual feedback, such as animations, during training.

\section{Distributed UMAP Experiments}
\label{sec:dist_experiments}

We tested the trustworthiness of our distributed UMAP implementation against the TASIC2018~\cite{tasic2018shared} dataset, which includes approx. 23k cells with 1k genes. Note that distributed UMAP performs inference only, as distributed training is an open problem. We trained a UMAP model on a single GPU using a random sample of the dataset and performed inference over partitions of the remaining data points. Section \ref{fig:umap_dist_viz} demonstrates that a reasonable trustworthiness can be a achieved by training on only 3\% of the dataset. Further, the increased variance when the number training samples decreases below 1\% appears to create the formation of more dense and tightly packed clusters. Still, we find a marginal impact to trustworthiness as the number of training samples is decreased and as the number of partitions is increased. 

\begin{figure}
    \centering
        \includegraphics[ width=\textwidth]{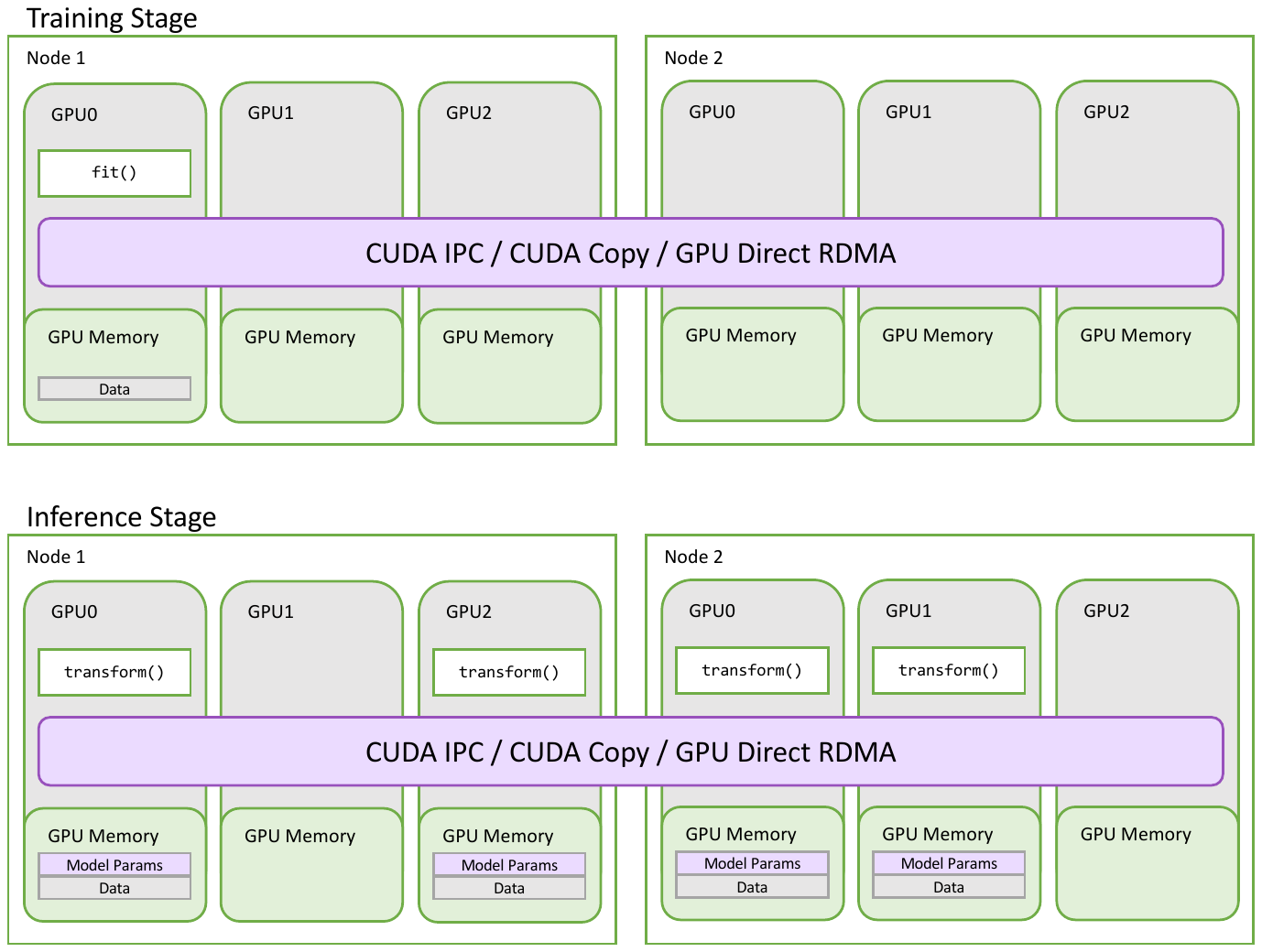}
    \caption{Distributed UMAP is executed on a cluster of workers, each mapped to a single GPU. A subsampling of the training data is used for training the model on a single worker and the model is scattered to workers containing data for out-of-sample prediction. UCX is used to transport GPU memory across the workers. }
    \label{fig:dist_umap}
\end{figure}

We executed performance tests for our distributed UMAP implementation against 10M randomly generated samples using Dask with 1, 2, 4, and 8 workers on a DGX1 containing 8x GV100 GPUs. For each experiment, we trained a UMAP model on 1\% of the data and started a timer. The trained model was broadcast to all of the workers in the Dask cluster and inference was performed in parallel. We stopped the timer when the data being inferenced was gathered back on the client. For UMAP-Learn experiments, we set the number of Numba threads to 80 and for \cuML{} experiments we mapped each worker to their own GPU. Section \ref{fig:dist_umap} contains the results of this experiment. While both \cuML{} and UMAP-learn achieve near-linear speedups as workers are added, \cuML{} dominated with a $255\times$ speedup on a single worker and  $100\times$ speedup on 8 workers. UMAP-learn would require 160 CPUs across 80 workers to achieve comparable performance.

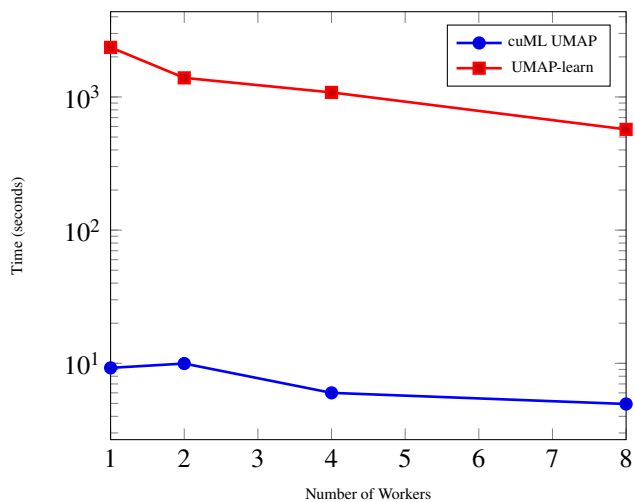
\begin{figure}[!htb]
\label{fig:umap_inference}
\begin{center}
\centering
\begin{tikzpicture}[]
\begin{axis}[
    xlabel={\tiny Number of Workers},
    ylabel={\tiny Time (seconds)},
    legend style={font=\tiny},
    legend pos=north east,
    legend columns=1,
    ymode=log,
    log basis y={10},
    xmin=1,
    xmax=8,
    ]
    
    \addplot+[line width=1.0pt] table [x=n_workers, y=gpu, col sep=comma] {mnmg_scaling.csv};
    \addlegendentry{\cuML}
    
    \addplot+[line width=1.0pt] table [x=n_workers, y=cpu, col sep=comma] {mnmg_scaling.csv};
    \addlegendentry{UMAP-learn}
    
\end{axis}

\end{tikzpicture}
\end{center}
\caption{Multi-GPU Scaling}
\label{fig:gpu_scale}
\end{figure}

\label{fig:umap_dist_viz}
\adjustbox{max width=\columnwidth}{%
\includegraphics[width=25cm, height=25cm]{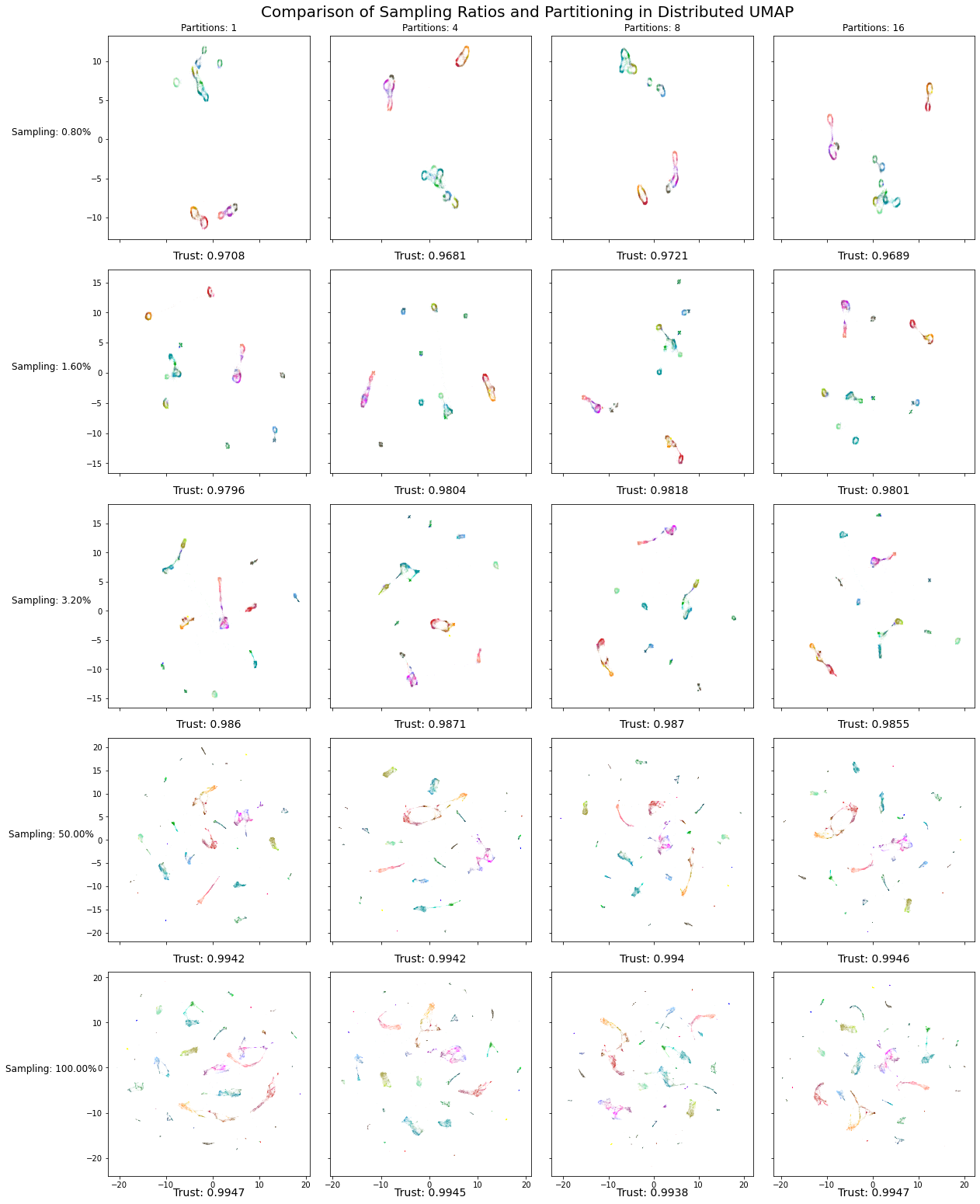}
}

\section{Experimenting with Neighborhood Sizes}

In addition to the unsupervised training experiments conducted with default values Table \ref{tbl:results_unsupervised}, we tested the three UMAP implementations with extreme values of $\mli{n\_neighbors}=5$ and $\mli{n\_neighbors}=50$. Following the experiments in Section \ref{sec:experiments}, these were also performed on a DGX1 containing $8\times$ 32gb V100 GPUs with $2\times$ Intel Xeon 20-core CPUs. 

\begin{table}[!h]
\caption{Each result shows mean $\pm$ variance, followed by max trustworhiness score, of each implementation of UMAP for the unsupervised case with $\mli{n\_neighbors}=5$. Fastest result in \textbf{bold}.}
\label{tbl:results_5_unsupervised}
\centering
\begin{tabular}{@{}lcccccc@{}}
\toprule
\multicolumn{1}{c}{} & \multicolumn{2}{c}{UMAP-Learn} & \multicolumn{2}{c}{GPUMAP}   & \multicolumn{2}{c}{\cuML}             \\ \cmidrule(lr){2-3} \cmidrule(lr){4-5} \cmidrule(l){6-7} 
Dataset              & $\mu \pm \sigma^2$    & Trust\%  & $\mu \pm \sigma^2$   & Trust\% & $\mu \pm \sigma^2$           & Trust\% \\ \midrule
digits               & $5.251 \pm 2.8944$  & 99.10    & $2.543 \pm 1.559$  & 96.69   & \textbf{0.3764$\pm$0.0109} & 99.23   \\
fashion mnist        & $29.82 \pm 2.7041$  & 98.19    & $3.932 \pm 1.913$  & 97.28   & \textbf{0.5432$\pm$0.0001} & 97.73   \\
mnist                & $33.63 \pm 0.7027$  & 96.30    & $5.029 \pm 1.872$  & 94.70   & \textbf{0.6712$\pm$0.0044} & 96.10   \\
cifar100             & $66.99 \pm 1.307$   & 86.87    & $4.984 \pm 1.861$  & 84.12   & \textbf{0.8252$\pm$0.0187} & 84.42   \\
coil20               & $9.384 \pm 0.001$   & 99.67    & $3.121 \pm 1.317$  & 96.23   & \textbf{0.3274$\pm$0.0178} & 99.44   \\
shuttle              & $29.88 \pm 5.204$   & 96.01    & $12.73 \pm 2.974$  & 93.29   & \textbf{0.6337$\pm$0.2727} & 96.80   \\ 
scRNA            & $161.22 \pm 6.435$   & 99.85       & $10.66 \pm 2.311$  & 99.88   & \textbf{3.8772$\pm$0.0108}  & 99.87   \\ \bottomrule
\end{tabular}
\end{table}

\begin{table}[!h]
\caption{Each result shows mean $\pm$ variance, followed by max trustworhiness score, of each implementation of UMAP for the unsupervised case with $\mli{n\_neighbors}=50$. Fastest result in \textbf{bold}.}
\label{tbl:results_50_unsupervised}
\centering
\begin{tabular}{@{}lcccccc@{}}
\toprule
\multicolumn{1}{c}{} & \multicolumn{2}{c}{UMAP-Learn} & \multicolumn{2}{c}{GPUMAP}   & \multicolumn{2}{c}{\cuML}             \\ \cmidrule(lr){2-3} \cmidrule(lr){4-5} \cmidrule(l){6-7} 
Dataset              & $\mu \pm \sigma^2$    & Trust\%  & $\mu \pm \sigma^2$   & Trust\% & $\mu \pm \sigma^2$           & Trust\% \\ \midrule
digits               & $7.922 \pm 3.0801$  & 98.01    & $2.423 \pm 1.433$  & 96.75   & \textbf{0.6380$\pm$0.5867} & 98.02   \\
fashion mnist        & $88.06 \pm 11.679$  & 96.69    & $8.428 \pm 1.895$  & 97.54   & \textbf{1.0485$\pm$0.0029} & 97.53   \\
mnist                & $119.96\pm 1.3089$  & 95.65    & $9.906 \pm 2.005$  & 95.46   & \textbf{1.0521$\pm$0.0022} & 95.27   \\
cifar100             & $222.52\pm 18.213$  & 84.20    & $13.08 \pm 3.214$  & 83.19   & \textbf{1.2524$\pm$0.0223} & 84.11   \\
coil20               & $12.364\pm 2.651$   & 97.40    & FAIL$\pm$FAIL  & FAIL  & \textbf{0.4217$\pm$0.0009} & 97.27   \\
shuttle              & $11.48 \pm 0.0156$  & 97.34    & FAIL$\pm$FAIL  & FAIL   & \textbf{0.3486$\pm$0.008} & 97.20   \\ 
scRNA            & $392.687 \pm 15.024$    & 69.49       & FAIL$\pm$FAIL  & FAIL   & \textbf{4.1645$\pm$0.008}  & 66.83   \\ \bottomrule
\end{tabular}
\end{table}

\section{Figures}
\label{sec:appending_figures}

\begin{algorithm}
\caption{The pairwise distance computations in our GPU-accelerated trustworthiness implementation are batched to preserve memory.}
\label{alg:strust}
\SetKwProg{scalabletrust}{Function \emph{ScalableTrust}}{}{end}
\scalabletrust{Matrix X, Matrix embed, Integer k, Integer n, Integer num\_batches}{
    \ForAll{$batch$ in num\_batches}{
      $nei\_orig = PairwiseDists(X[batch,:])$\;
     $nei\_embed = KNearestNeighbors(embed, k)$\;
     $t = 0$\;
     \ForAll{row $c$ in $nei\_orig$}{
        $t = t + Rank(neigh\_orig, nei\_embed, k)$\;
     }
     }
     return $t(1 - k(2 / (nk * 2n - 3k) - 1))$\;
}
\end{algorithm}


\begin{figure}
\centering
\includegraphics[trim=0 0 0 20, clip, width=13cm, height=7cm]{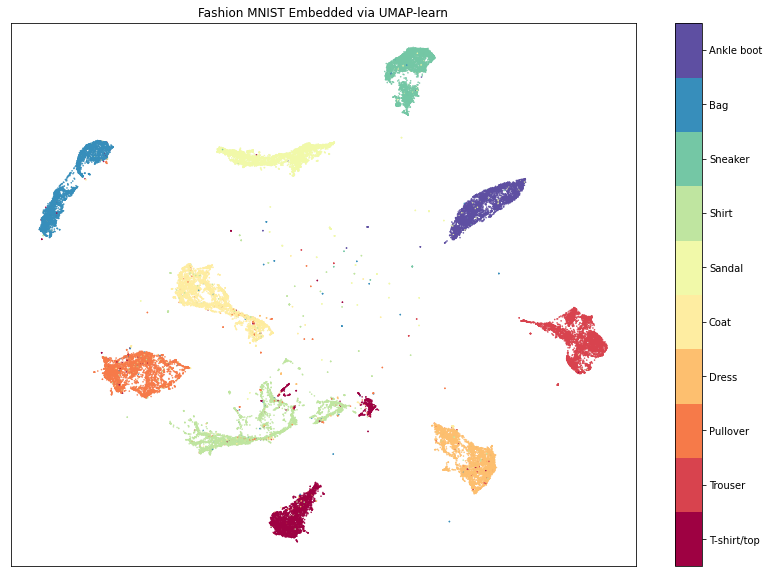}
\caption{Fashion MNIST embedded with UMAP-Learn}
\end{figure}

\begin{figure}
\centering
\includegraphics[trim=0 0 0 20, clip, width=13cm, height=7cm]{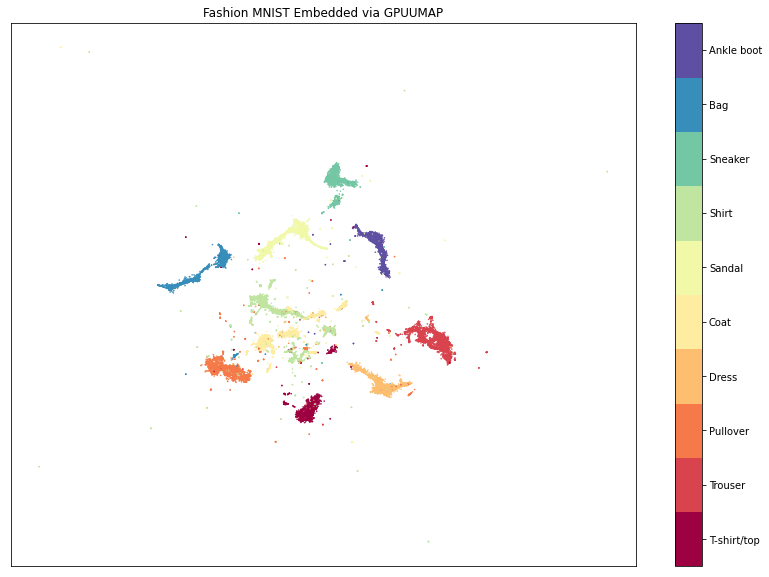}
\caption{Fashion MNIST embedded with GPUMAP}
\end{figure}

\begin{figure}
\centering
\includegraphics[trim=0 0 0 20,clip, width=13cm, height=7cm]{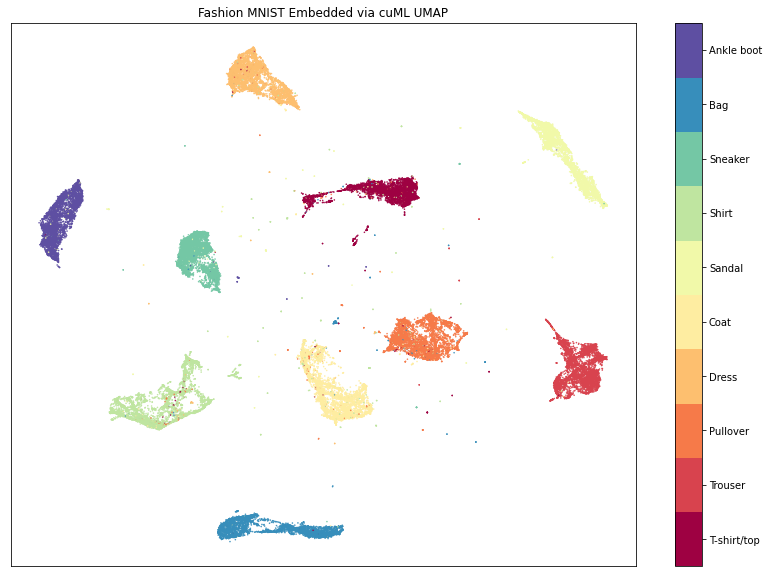}
\caption{Fashion MNIST embedded with \cuML}
\end{figure}

\begin{figure}
    \centering
        \includegraphics[width=\textwidth]{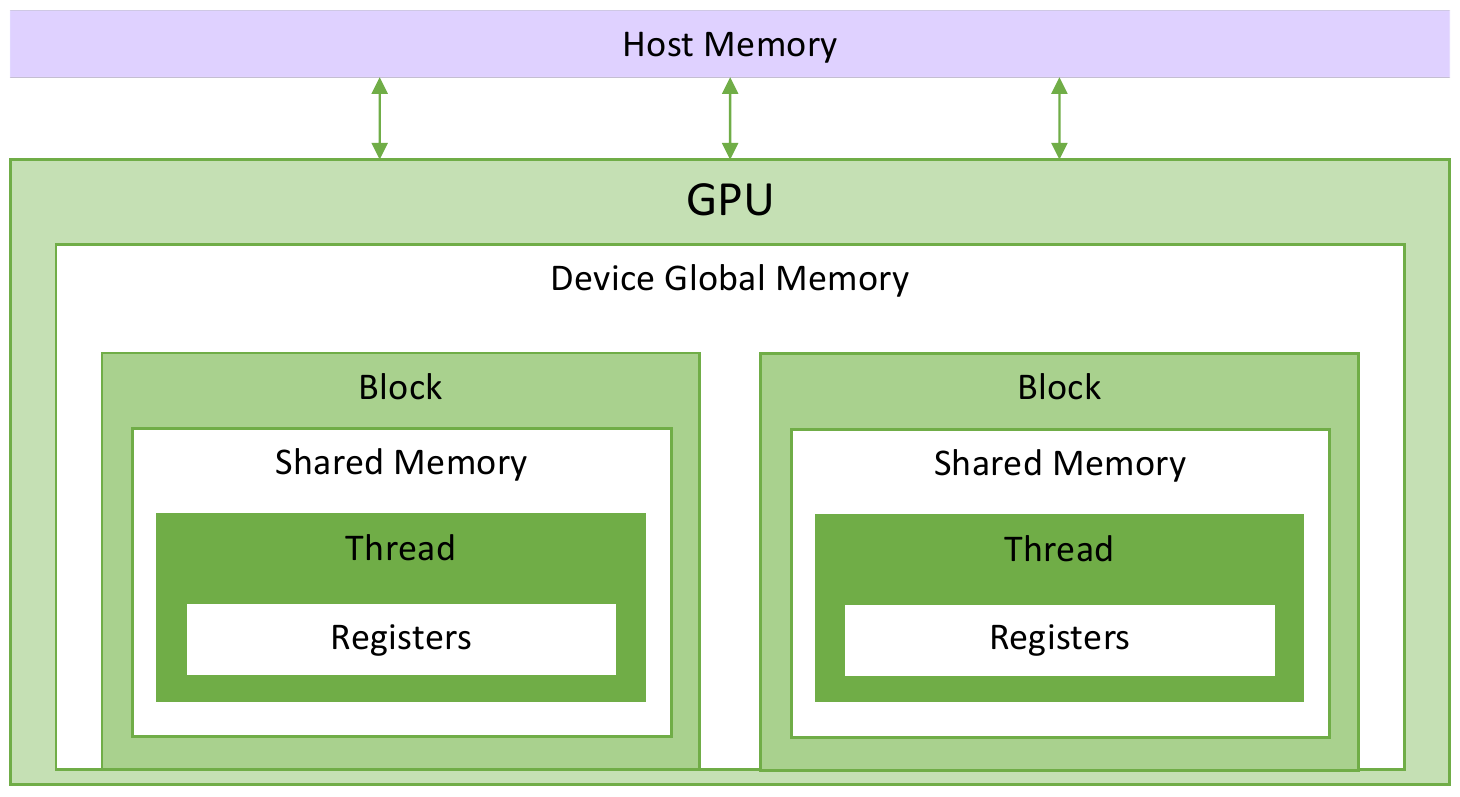}
    \caption{The GPU Architecture contains global device memory that is accessible by several thread-blocks. Each thread-block contains shared memory which can be accessed by their internal threads. Threads each contain a set of registers.}
    \label{fig:gpu_arch}
\end{figure}

\end{document}